\documentclass[letterpaper]{article} 
\usepackage{aaai24}  
\usepackage{times}  
\usepackage{helvet}  
\usepackage{courier}  
\usepackage[hyphens]{url}  
\usepackage{graphicx} 
\usepackage{natbib}  
\usepackage{caption} 
\usepackage{algorithm}
\usepackage{algorithmic}
\usepackage{amsmath}
\usepackage{amssymb}
\usepackage{amsbsy}
\usepackage{subfig}
\usepackage{bm}
\usepackage{multirow}
\usepackage{mathtools}

\usepackage{newfloat}
\usepackage{listings}
\DeclareCaptionStyle{ruled}{labelfont=normalfont,labelsep=colon,strut=off} 
\floatstyle{ruled}
\newfloat{listing}{tb}{lst}{}
\floatname{listing}{Listing}
\nocopyright

\title{STDiff: Spatio-temporal Diffusion for Continuous Stochastic Video Prediction}
\author{
    Xi Ye\textsuperscript{\rm}\thanks{Corresponding Author. \\Accepted by AAAI 2024.}, Guillaume-Alexandre Bilodeau
}
\affiliations{
    LITIV, Polytechnique Montréal\\

    xi.ye@polymtl.ca, gabilodeau@polymtl.ca
}

\usepackage{bibentry}

\begin{document}

\maketitle

\begin{abstract}
Predicting future frames of a video is challenging because it is difficult to learn the uncertainty of the underlying factors influencing their contents. In this paper, we propose a novel video prediction model, which has infinite-dimensional latent variables over the spatio-temporal domain. Specifically, we first decompose the video motion and content information, then take a neural stochastic differential equation to predict the temporal motion information, and finally, an image diffusion model autoregressively generates the video frame by conditioning on the predicted motion feature and the previous frame. The better expressiveness and stronger stochasticity learning capability of our model lead to state-of-the-art video prediction performances. As well, our model is able to achieve temporal continuous prediction, i.e., predicting in an unsupervised way the future video frames with an arbitrarily high frame rate. Our code is available at \url{https://github.com/XiYe20/STDiffProject}.
\end{abstract}

\section{Introduction}

\label{sec:intro}
Given some observed past frames as input, a video prediction model forecasts plausible future frames, with the aim of mimicking the vision-based precognition ability of humans. Anticipating the future is critical for developing intelligent agents, thus video prediction models have applications in the field of autonomous driving, route planning, and anomaly detection \cite{lu2019}. 

Video frame prediction (VFP) is challenging due to the inherent unpredictability of the future. Theoretically, there are an infinite number of potential future outcomes corresponding to the same past observation. Moreover, the stochasticity increases exponentially as the model predicts towards a more distant future. Early deterministic video prediction models \cite{finn2016, villegas2017a} are incapable of dealing with a multimodal future and thus the prediction tends to be blurry. Subsequently, techniques such as VAE and GANs were introduced for stochastic video prediction \cite{babaeizadeh2018, lee2018, denton2018, castrejon2019}. However, they still fall short in achieving satisfactory results.

Previous works \cite{castrejon2019, wu2021a} empirically observed that the main issue which limits the performance is that the stochastic video prediction model is not expressive enough, specifically, shallow levels of latent variables are incapable of capturing the complex ground-truth latent distribution. \citet{castrejon2019} introduced hierarchical latent variables into variational RNNs (VRNNs) to mitigate the problem. Recent progress of diffusion-based video prediction models \cite{voleti2022, harvey2022, hoppe2022} can also be attributed to the increase of model expressiveness, because diffusion models can be considered as infinitely deep hierarchical VAEs \cite{huang2021, tzen2019} with a fixed encoder and latent variables across different levels having the same dimensionality as the data.

However, none of the previous models explored extending the stochastic expressiveness along the temporal dimension. Hierarchical VRNNs incorporate randomness at each fixed time step, which is limited by the discrete nature of RNNs. Meanwhile, almost all the previous diffusion-based video prediction models \cite{voleti2022, harvey2022, hoppe2022} concatenate a few frames together and learn the distribution of those short video clips as a whole, which ignores an explicit temporal stochasticity estimation between frames. Therefore, there is a need  to increase the expressiveness of the stochastic video prediction model over both the spatial and temporal dimensions.

One more issue is that almost all video prediction models can only predict future frames at fixed time steps, ignoring the continuous nature of real-world dynamic scenes. Thus, additional video interpolation models are required to generate videos with a different frame rate. Only a few recent preliminary works have explored this topic \cite{park2021b, ye2023}, but they are either limited to deterministic predictions \cite{park2021b} or exhibit low diversity in stochastic predictions \cite{ye2023}. To improve video prediction, a method should be able to do continuous stochastic prediction with a good level of diversity.

In this paper, we propose a novel diffusion-based stochastic video prediction model that addresses the limited expressiveness and the temporal continuous prediction problems for the video prediction task. We propose to increase the expressiveness of the video prediction model by separately learning the temporal and spatial stochasticity. Specifically, we take the difference images of adjacent past frames as motion information, and those images are fed into a Conv-GRU to extract the initial motion feature of future frames. Given the initial motion feature, a neural stochastic differential equation (SDE) solver predicts the motion feature at an arbitrary future time, which enables continuous temporal prediction. Finally, an image diffusion model conditions on the previous frame and on the motion feature to generate the current frame. Because the diffusion process can also be described by SDE \cite{song2021}, our model explicitly uses SDEs to describe both the spatial and temporal latent variables, and thus our model is more flexible and expressive than previous ones. Our contributions can be summarized as follows:

\begin{itemize}
\item We propose a novel video prediction model with better expressiveness by describing both the spatial and temporal generative process with SDEs;

\item Our model attains state-of-the-art (SOTA) FVD and LPIPS score across multiple video prediction datasets;

\item Compared to prior approaches, our model significantly enhances the efficiency of generating diverse stochastic predictions;

\item To the best of our knowledge, this is the first diffusion-based video prediction model with temporal continuous prediction and with motion content decomposition.

\end{itemize}

\section{Related works}

Video prediction models can be classified into two types:  deterministic or stochastic. Since we aim at addressing uncertainty, we focus on the stochastic models in this section.
Most stochastic video prediction models \cite{villegas2017a, babaeizadeh2018, denton2018, lee2018} utilize a variational RNN (VRNN) as the backbone. The performance of VRNN-based models is constrained as there is only one level of latent variables. Hierarchical VAEs are utilized by some works \cite{castrejon2019, chatterjee2021, wu2021a} to deal with the stochasticity underfitting problem. FitVid \cite{babaeizadeh2021} proposed a carefully designed VRNN architecture to address the underfitting problem.  Despite the improvements, the stochasticity of all these models is still characterized by the variance of the spatial latent variables of each frame \cite{chatterjee2021}. 

Some methods, such as MCNet \cite{villegas2017a}, decompose the motion and content of videos with the assumption that different video frames share similar content but with different motion, thus the decomposition of motion and content facilitates the learning. It was applied only spatially. Given its benefit, this strategy inspired us to learn the stochasticity over the temporal and spatial dimension separately. NUQ \cite{chatterjee2021} is the only VRNN-based model that considers the temporal predictive uncertainty. It enforces an uncertainty regularization to the vanilla loss function for a better convergence without modification of the VRNN architecture. In contrast, our proposed model employ an SDE to explicitly account for the randomness of the temporal motion.

Given the success of image diffusion models, some works have adapted them for video prediction \cite{hoppe2022, voleti2022, nikankin2022, harvey2022, yang2022}. Almost all diffusion-based models are non-autoregressive models that learn to estimate the distribution of a short future video clip. Therefore, they ignore the explicit learning of motion stochasticity and generate multiple future frames with a fixed frame rate all at once. \citet{yang2022} combined a deterministic autoregressive video prediction model with a diffusion model, where the latter is used to generate stochastic residual change between frames to correct the deterministic prediction. TimeGrad \cite{rasul2021} combines an autoregressive model with a diffusion model, but it is only for low dimensional time series data prediction. In contrast, our method combines an autoregressive model with a diffusion model for videos and can make a continuous temporal prediction.

Besides, few recent works investigated continuous video prediction, including Vid-ODE \cite{park2021b} and NPVP \cite{ye2023}. Vid-ODE \cite{park2021b} combines neural ODE and Conv-GRU to predict the features of future frames and a CNN decoder is taken to decode the frames in a compositional manner. It achieves temporal continuous predictions, but it is deterministic.  NPVP \cite{ye2023} is a non-autoregressive model that tackles the continuous problem by formulating the video prediction as an attentive neural process. However, the diversity of the stochastic predictions is low because NPVP only takes a single global latent variable for the whole sequence to account for the stochasticity. Our use of SDEs allows us to obtain better diversity.

\section{Background}
\subsection{Neural SDEs}
Stochastic differential equations are widely used to describe the dynamics in engineering. Compared with an ordinary differential equation (ODE), a SDE takes into account stochasticity by incorporating randomness into the differential equation. Given an observation $X_t$ at time $t$, an Itô SDE is formulated as 
\begin{equation}
    dX_t = \mu(X_t, t)dt + \sigma(X_t, t)dW_t,
\label{eq:sde}
\end{equation}

\noindent where $W$ denotes a Wiener process (Brownian motion). The first term and second term of the right hand side are the drift term and diffusion term respectively. We can integrate Eq. \ref{eq:sde} by Itô's calculus. In this paper, we use a simpler version of SDE, where $\mu$ and $\sigma$ are independent of $t$.

Given observations $X_t$, we can fit an SDE by parameterizing $\mu$ and $\sigma$ with neural networks. This is referred to as neural SDEs \cite{li2020a, kidger2021}. \citet{li2020a} generalized the adjoint sensitivity method for ODEs into SDEs for an efficient gradient computation to enable the learning of neural SDEs. 

\subsection{Diffusion models}
\label{ssec:diffusion_models}

There are two types of equivalent diffusion models. The first one is denoising diffusion probabilistic models (DDPM) \cite{ho2020, sohl-dickstein2015}. DDPM consists of a fixed forward diffusion and a learned reversed denoising process. The forward diffusion process gradually adds noise to the data until it converges to a standard Gaussian. During the reverse process (generative process), a random Gaussian noise is sampled, and a trained neural network is repeatedly applied to denoise and finally converts the random Gaussian noise to a clean data sample. The second type is the score matching-based method \cite{song2021}, which describes both the forward and reversed diffusion processes as SDEs. \citet{song2021} studied the connection between the two methods and proved that a DDPM is equivalent to a discretization of a score matching model based on continuous variance preserving SDEs (VP-SDEs). As we also model the motion feature of videos by a SDE in this paper, we present the image diffusion model under the framework of SDEs, specifically, the VP-SDE score matching model.

The forward and reversed processes of the VP-SDE score matching model are given in Eq. \ref{eq:f_vpsde} and Eq. \ref{eq:r_vpsde} respectively:
\begin{equation}
dx_t = -\frac{1}{2}\beta(t)x_tdt + \sqrt{\beta(t)}dW_t \label{eq:f_vpsde}
\end{equation}
\begin{equation}
dx_t = [-\frac{1}{2}\beta(t)x_t - \beta(t)\nabla_{x_t}\text{log}q_t(x_t)]dt + \sqrt{\beta(t)}d\bar{W}_t, \label{eq:r_vpsde}
\end{equation}

\noindent where $t\sim U[0, T]$ denotes a random diffusion timestep with a maximum value of $T$, $x_T\sim \mathcal{N}(0, I)$, $x_t\sim q_t(x_t)$ is the perturbed image at diffusion step $t$, $\beta(t)$ denotes the forward noise schedule, and $\bar{W}_t$ denotes a backward Wiener process. The VP-SDE score matching model is trained by minimizing the following score matching loss:
\begin{align}
    &E_{t}E_{x_0}E_{x_t\sim q_t(x_t|x_0)}\left\lVert \bm{s}_\theta(x_t, t) - \nabla_{x_t}\text{log}q_t(x_t|x_0)\right\rVert^2,
\label{eq:score_match_loss}
\end{align}

\noindent where $\bm{s}_\theta$ denotes the score estimation neural network, and $q_t(x_t|x_0)=\mathcal{N}(x_t;\gamma_t x_0, \sigma_t^2 I)$ denotes the distribution of perturbed image $x_t$ with $\gamma_t = e^{-\frac{1}{2}\int_0^t\beta(s)ds}$ and $\sigma_t^2=1-e^{-\int_0^t\beta(s)ds}$. Therefore, the forward diffusion process described by Eq. \ref{eq:f_vpsde} can be achieved by direct re-parameterized sampling, i.e., $x_t=\gamma_t x_0 + \sigma_t \epsilon$, where $\epsilon \sim \mathcal{N}(0, I)$. In this case, we can parameterize $\bm{s}_\theta(x_t,t)$ by $-\frac{\bm{\epsilon}_\theta(x_t,t)}{\sigma_t}$, where $\bm{\epsilon}_\theta$ denotes a noise predictor neural network. Then the score matching loss in Eq. \ref{eq:score_match_loss} can be reduced to an equivalent noise prediction loss as DDPM:
\begin{equation}
E_{t}E_{x_0}E_{\epsilon\sim \mathcal{N}(0,1)}\frac{1}{\sigma_t^2}\left\lVert \epsilon - \bm{\epsilon}_\theta(x_t, t)\right\rVert^2,
\label{eq:ddpm_loss}
\end{equation}

\noindent except that here, the diffusion timestep $t$ is continuous instead of discrete. Finally, we can apply loss weighting $\lambda(t) = \sigma_t^2$ to Eq. \ref{eq:ddpm_loss} for a good perceptual quality, which also further simplifies the loss function to be:
\begin{equation}
E_{t}E_{x_0}E_{\epsilon\sim \mathcal{N}(0,1)}\left\lVert \epsilon - \bm{\epsilon}_\theta(x_t, t)\right\rVert^2.
\label{eq:ddpm_loss_simple}
\end{equation}

Under the variational framework, \citet{huang2021} proved that continuous-time diffusion models, like the VP-SDE score matching model, can be viewed as ``the continuous limit of hierarchical VAEs'' as purposed by \citet{tzen2019}. In other words, a continuous diffusion model is equivalent to an infinitely deep hierarchical VAE. 

\section{Methodology}
\begin{figure}[ht]
\centering
\includegraphics[width=0.6\linewidth]{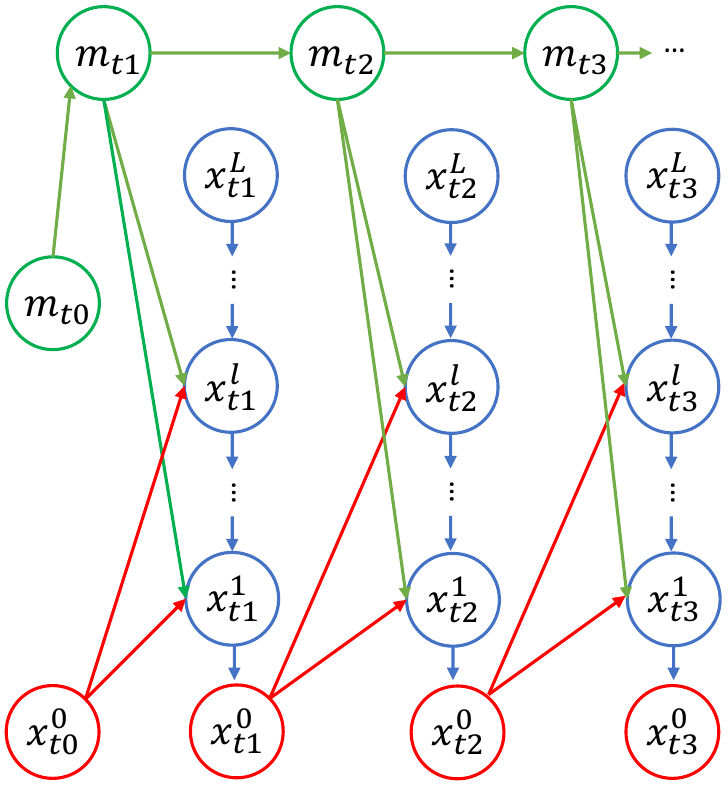}
\caption{Graphical model for the generation process of STDiff. Green arrows denote the temporal motion connections, and blue arrows denote the connections between latent variables of each frame at timestep $t$, i.e., the reverse image diffusion process. Red arrows denote the recurrent connection from the previous frame to each level of latent variable in the next time step. $m_{t_0}$ denotes the initial motion feature extracted from observed frames. $x_{t_0}^0$ denotes the most recent observed frame.}
\label{fig:STDiff_graph}
\end{figure}

Note that in this section, $t$ is used to denote video temporal coordinates. We address the video prediction problem that predicts $P$ future frames $\bm{x}=\{x^0_{t_1}, x^0_{t_2}, ..., x^0_{t_P}\}$ given $N$ past frames $\bm{c}= \{x^0_1, x^0_2, ..., x^0_N\}$. During training, $t_i$ is a discrete future frame integer index. However, $t_i$ can be an arbitrary positive real number timestep during inference. The training objective is to maximize $p(\boldsymbol{x}|\boldsymbol{c})$.  In order to simplify the learning problem, we decompose the motion and content features and we also factorize the joint distribution along the temporal dimension, i.e., autoregressive prediction.

\begin{figure*}[t]
\centering
\includegraphics[width=0.65\linewidth]{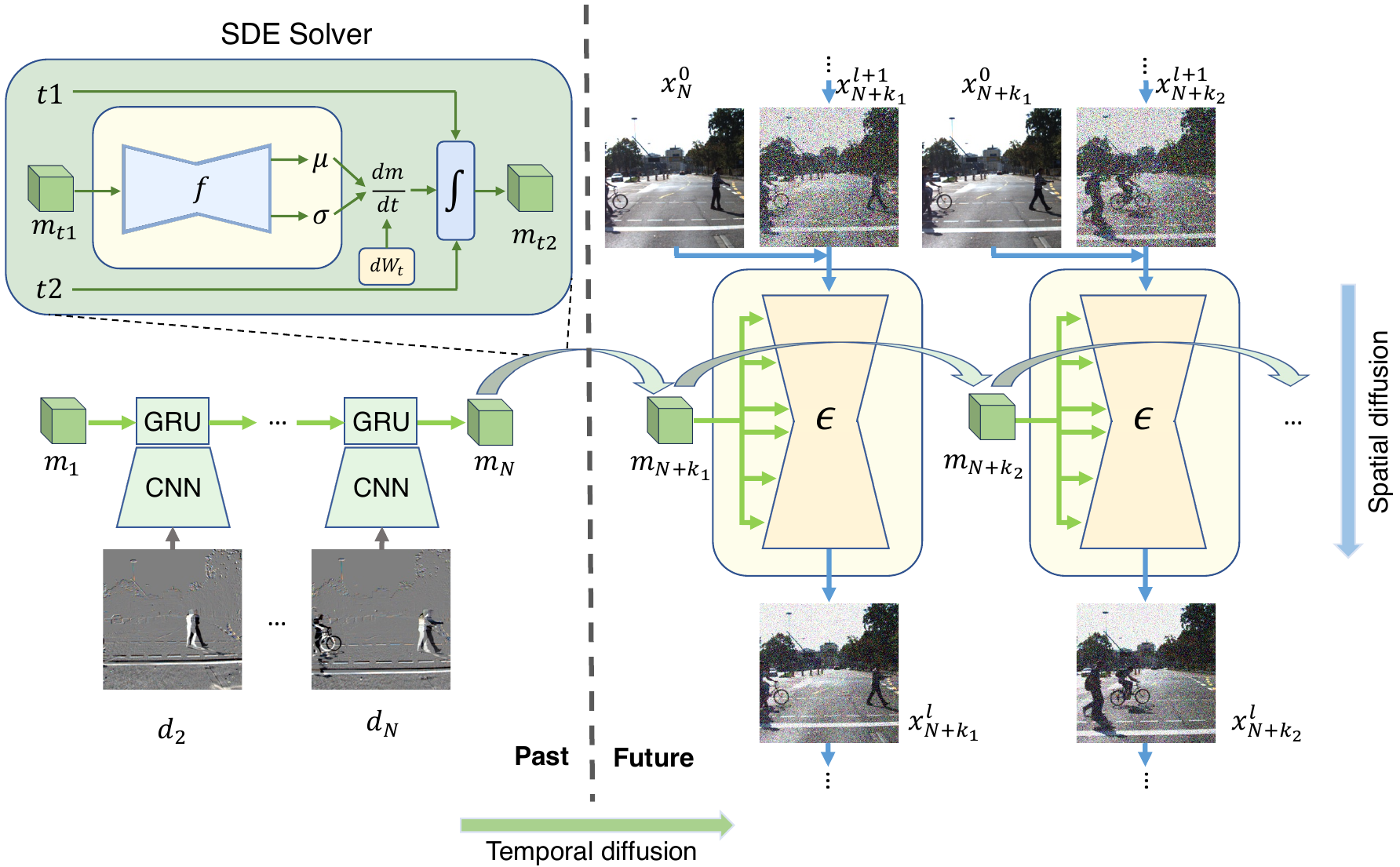}

\caption{Neural network architecture of STDiff. Difference images of past frames are encoded as motion feature $m_N$ by a ConvGRU for future motion prediction. The curved arrow denotes one step of random future motion prediction (SDE integration), i.e., the temporal motion diffusion process. Detail computation flow of the SDE solver is shown in the top left green box. The conditional image diffusion model recurrently predicts each future frame given motion feature and the previous frame.}
\label{fig:STDiff_model}
\end{figure*}

Considering the close relationship between diffusion models and hierarchical VAEs \cite{huang2021}, we formulate our generative probabilistic model under the framework of VRNNs (Figure \ref{fig:STDiff_graph}) as an extended hierarchical recurrent RNN, but with infinite-dimensional latent variables over the spatio-temporal domain. To avoid confusion, we use $l$ instead of $t$ to denote the spatial diffusion timesteps. $l=0$ denotes the pixel space. Given the initial motion feature $m_{t_0}$ extracted from $\bm{c}$ and the most recent observed frame $x^0_{t_0}$, i.e., $x^0_{N}$, we can first factorize the probability as follows with the assumption that the temporal process is Markovian:
\begin{align}
    p(\boldsymbol{x}|\boldsymbol{c}) = p(\boldsymbol{x}|x^0_{t_0}, m_{t_0}) = \prod_{i=1}^{P}p(x_{t_i}^0|x^0_{t_{i-1}}, m_{t_{i-1}}),
\end{align}
where the motion feature $m_{t_i}$ follows a transitional distribution $p(m_{t_i}|m_{t_{i-1}})$ defined by the temporal neural SDE in Eq.\ref{eq:motion_sde}. Following the same technique of conditional DDPM \cite{ho2020}, we can derive the following learning objective:

\begin{equation}
\mathcal{L}_\theta = \sum_{i=1}^{P}\mathbb{E}_q[\mathcal{L}_L + \sum_{l>1}\mathcal{L}_{l-1} + \mathcal{L}_0],
\label{eq:stdiff_loss}
\end{equation}
where each term is defined as follows:
\begin{equation}
\mathcal{L}_L = D_{KL}(q(x_{t_i}^L|x_{t_i}^0)||p_\theta(x_{t_i}^L)),
\end{equation}
\begin{equation}
\mathcal{L}_{l-1} = D_{KL}(q(x_{t_i}^{l-1}|x_{t_i}^l, 
x_{t_i}^0)||p_\theta(x_{t_i}^{l-1}|x_{t_i}^l, x_{t_{i-1}}^0, m_{t_i})),
\end{equation}
\begin{equation}
\mathcal{L}_0 = - \log p_\theta(x_{t_i}^0|x_{t_i}^1, x_{t_{i-1}}^0, m_{t_i})
\end{equation}

By applying the techniques described in \cite{ho2020}, we can simplify the objective in Eq. \ref{eq:stdiff_loss} to be 
\begin{equation}
\mathcal{L}_\theta = \sum_{i=1}^{P}E_{l}E_{x_{t_i}^0}E_{\epsilon\sim \mathcal{N}(0,1)}\left\lVert \epsilon - \bm{\epsilon}_\theta(x_{t_i}^l, x^0_{t_{i-1}}, m_{t_i}, l)\right\rVert^2.
\end{equation}
Please refer to the appendix for the detailed derivation of the loss function.

\subsection{Spatio-Temporal Diffusion (STDiff) architecture}
We call our proposed method STDiff (Spatio-Temporal Diffusion) and its architecture is shown in Figure \ref{fig:STDiff_model}. It consists of two parts, a motion predictor and a recurrent diffusion frame predictor. The motion predictor encodes all the past motion features and predicts future motion features. Given the predicted motion feature at different future time steps and the previous frame, the recurrent diffusion frame predictor generates the desired future frame. The detailed architecture of the two modules is described as follows.

\noindent \textbf{Motion Predictor.} We decompose the motion and content feature because the SDEs are naturally designed for dynamic information modeling, and also it eases the learning burden of the neural SDE. The motion predictor is divided into two parts: 1) a Conv-GRU for past motion encoding and 2) a neural SDE for future motion prediction. Assuming a regular temporal step in the observed past frames, we utilize a Conv-GRU for past motion encoding due to its flexibility and efficiency. In order to achieve the motion and appearance decomposition, we calculate the difference images $d_i$ of adjacent past frames as the input of the Conv-GRU. Given the zero-initialized motion hidden state $m_1$ and $N-1$ difference images, the Conv-GRU outputs the motion feature $m_N$ for a future prediction.

\begin{table*}[t]
\setlength{\tabcolsep}{2pt}
\centering
\resizebox{\textwidth}{!}{
\begin{tabular}{lccc} \hline
\multirow{2}{*}{Models} & \multicolumn{3}{c}{SMMNIST, \textit{5 $\rightarrow$ 10}}\\
& FVD$\downarrow$ & SSIM$\uparrow$ & LPIPS$\downarrow$ \\ 
\hline
SVG-LP \cite{denton2018} & 90.81 & 0.688 & 153.0 \\
Hier-VRNN \cite{castrejon2019}& 57.17 & 0.760 & \textbf{103.0}\\
MCVD-concat \cite{voleti2022} & 25.63 & \underline{0.786} & - \\
MCVD-spatin \cite{voleti2022} & \underline{23.86} & 0.780 & - \\
NPVP \cite{ye2023} & 95.69 & \textbf{0.817} & 188.7 \\ \hline
\textit{STDiff} (ours)& \textbf{14.93} & 0.748 & \underline{146.2}\\
\hline
\end{tabular}

\quad
\begin{tabular}{lccc} \hline
\multirow{2}{*}{Models} & \multicolumn{3}{c}{BAIR, \textit{2 $\rightarrow$ 28}} \\
& FVD$\downarrow$ & SSIM$\uparrow$ & LPIPS$\downarrow$\\ 
\hline

SAVP \cite{lee2018} & 116.4 & 0.789 & 63.4 \\ 
Hier-VRNN \cite{castrejon2019} & 143.4 & 0.829 & \textbf{55.0}\\ 
STMFANet \cite{jin2020} & 159.6 & \textbf{0.844} & 93.6\\
VPTR-NAR \cite{ye2022} & - & 0.813 & 70.0 \\
NPVP \cite{ye2023} & 923.62 & \underline{0.842} & \underline{57.43}\\
MCVD-concat \cite{voleti2022} &120.6 & 0.785 & 70.74 \\
MCVD-spatin \cite{voleti2022} &132.1 & 0.779 & 75.27 \\
FitVid \cite{babaeizadeh2021} & \underline{93.6} & - & - \\
\hline
\textit{STDiff} (ours) & \textbf{88.1} & 0.818 & 69.40
  \\ \hline
\end{tabular}}

\caption{VFP results on SMMNIST (left) and BAIR (right). \textbf{Boldface}: best results. \underline{Underlined}: second best results.}
\label{tab:SMMNIST_VFP}

\end{table*}

Given $m_N$ as the initial value of future motion features, we fit the future motion dynamic by a neural SDE, which is equivalent to a learned diffusion process. Specifically, given the motion feature $m_{t_{i-1}}$ at time step $t_{i-1}$, a small neural network $f_\theta$ is taken to parameterize the drift coefficient and diffusion coefficient respectively. Then, the motion feature at $t_i$ is integrated with
\begin{align}
\mu, \sigma &= f_\theta(m_{t_{i-1}}) \\
m_{t_i} &= m_{t_{i-1}} + \int_{t_{i-1}}^{t_i}\mu dt + \int_{t_{i-1}}^{t_i}\sigma dW_t.
\label{eq:motion_sde}
\end{align}
In Figure \ref{fig:STDiff_model}, each curved arrow denotes one future motion prediction step, i.e., one integration step for the neural SDE. For better learning of the temporal dynamics, we randomly sample $t_i$ from the future time steps during training (see Algorithm \ref{alg:Training} for details about the training).

\noindent \textbf{Recurrent diffusion predictor.} At a future time step $t_i$, given a noisy frame $x_{t_i}^l$ derived from the forward diffusion process, a UNet is trained to predict the noise condition on the previous clean frame $x_{t_{i-1}}^0$ and the motion feature $m_{t_i}$. In detail, $x_{t_{i-1}}^0$ is concatenated with $x_{t_i}^l$ as the input of the UNet, and $m_{t_i}$ is fed into each ResNet block of the UNet by the manner of spatially-adaptive denormalization \cite{park2019}. Please see Algorithm \ref{alg:Training} for training details.

The inference process of the recurrent diffusion predictor is almost the same as the prediction process, with the distinction that the model generates the previous clean frame $x_{t_{i-1}}^0$ during the previous time step, and subsequently feeds it back into the UNet input for forecasting the next frame.

\begin{algorithm}
\caption{Training}\label{alg:Training}
\hspace*{\algorithmicindent} \textbf{Input} Observed frames: $V_o=\{x_1, x_2, ...,x_N\}$ \\ 
\hspace*{\algorithmicindent} Frames to predict: $V_p=\{x_{N+1}, x_{N+2}, ..., x_{N+M}\}$ \\
\hspace*{\algorithmicindent} \textbf{Initialize} $\text{ConvGRU}_\theta$ , $f_\theta$, $\epsilon_\theta$, 
\begin{algorithmic}[1]

\REPEAT

\STATE Initialize motion states: $m_1 = \bm{0}$
\FOR{$i=2, ..., N$}
\STATE $d_i = x_i-x_{i-1}$
\STATE $m_i = \text{ConvGRU}_\theta(d_i, m_{i-1})$
\ENDFOR

\STATE $\{k_1,..,k_P\}=\text{Sort}(\text{RandomChoice}(\{1, ..., M\})$)
\STATE $\{t_0, t_1, ..., t_P\}$ = \{$N, N+k_1, ..., N+k_P$\}
\FOR{$i=1, ..., P$}
\STATE $m_{t_i} = \text{SDESolver}(f_\theta, m_{t_{i-1}}, (t_{i-1}, t_i))$
\STATE $l \sim Uniform(0, L)$, $\epsilon \sim \mathcal{N}(\textbf{0}, \textbf{I})$
\STATE $x_{t_i}^l = \gamma_l x_{t_i}^0 + \sigma_l\epsilon$
\STATE \text{Take gradient descent step on}
\STATE \hspace{\algorithmicindent} $\nabla_\theta \lVert \epsilon - \epsilon_\theta(x_{t_i}^l, x^0_{t_{i-1}}, m_{t_i}, l)\rVert^2$
\ENDFOR
\UNTIL convergence
\end{algorithmic}
\end{algorithm}

\section{Experiments}

\begin{table*}[!ht]
\setlength{\tabcolsep}{1pt}
\centering
\resizebox{\textwidth}{!}{
\begin{tabular}{lccc} \hline
\multirow{2}{*}{Models} & \multicolumn{3}{c}{KITTI, \textit{4 $\rightarrow$ 5}} \\
& FVD $\downarrow$ & SSIM$\uparrow$ & LPIPS$\downarrow$ \\ \hline
PredNet \cite{lotter2017} & - & 0.48 & 629.5\\
Voxel Flow \cite{liu2017} & - & 0.43 & 415.9\\
SADM \cite{bei2021} & - & 0.65 & 311.6 \\
Wu et al. \cite{wu2022} & - & 0.61 & 263.5 \\
DMVFN \cite{hu2023b} & - & \textbf{0.71} & \underline{260.5} \\
NPVP \cite{ye2023} & \underline{134.69} & \underline{0.66} & 279.0 \\ \hline
\textit{STDiff} (ours) & \textbf{51.39} & 0.54 & \textbf{114.6} \\ \hline
\end{tabular}
\quad
\begin{tabular}{lccc} \hline
\multirow{2}{*}{Models} & \multicolumn{3}{c}{Cityscapes, \textit{2 $\rightarrow$ 28}} \\
& FVD$\downarrow$ & SSIM$\uparrow$ & LPIPS$\downarrow$ \\ 
\hline

SVG-LP \cite{denton2018} & 1300.26 & 0.574 & 549.0 \\ 
VRNN 1L\cite{castrejon2019} & 682.08 & 0.609 & 304.0 \\
Hier-VRNN \cite{castrejon2019}& 567.51 & 0.628 & 264.0 \\
GHVAEs \cite{wu2021a}& 418.00& \underline{0.740} & 194.0 \\
MCVD-concat \cite{voleti2022} & \underline{141.31} & 0.690 & \textbf{112.0} \\
NPVP \cite{ye2023} & 768.04 & \textbf{0.744} & 183.2 \\
\hline
\textit{STDiff} (ours) & \textbf{107.31} & 0.658 & \underline{136.26}\\
\hline
\end{tabular}}

\caption{VFP results on KITTI (left) and Cityscapes (right). \textbf{Boldface}: best results. \underline{Underlined}: second best results.}
\label{tab:KTH_CityScapes_KITTI_VFP_results}
\end{table*}

We evaluated the performance of the proposed STDiff model on KITTI \cite{geiger2013}, Cityscapes \cite{cordts2016}, KTH \cite{schuldt2004}, BAIR \cite{ebert2017} and stochastic moving MNIST (SMMNIST) \cite{denton2018} datasets. KITTI and Cityscapes are high-resolution traffic video datasets used to evaluate our model real-world application capabilities. The KTH dataset consists of different human motion videos, BAIR includes videos of a randomly moving robot arm that pushes different objects. BAIR and SMMNIST pose greater challenges due to their higher levels of stochasticity compared to the three others. The number of past frames and future frames to predict is determined according to the experimental protocols of previous works (see the supplementary material for more details). All models are trained with 4 NVIDIA V100 Volta GPU (32G memory).

\subsection{Frame prediction performance}

We firstly tested STDiff given the same protocol as previous work, i.e., report the best SSIM, LPIPS out of multiple different random predictions, together with the FVD score. Note that we only sample 10 different random predictions for each test example as for MCVD, instead of 100 different predictions for all other previous stochastic models. The results are presented in Table \ref{tab:SMMNIST_VFP} and Table \ref{tab:KTH_CityScapes_KITTI_VFP_results}.

In Table \ref{tab:SMMNIST_VFP}, we observe that STDiff achieves the new SOTA for the FVD score on SMMNIST, and also the second best LPIPS score. For BAIR, STDiff also outperforms all previous work in terms of FVD score. The results show that the random predictions of STDiff have a better temporal coherence and match the ground-truth distribution much better, i.e., the predictions are more natural and plausible.

\begin{figure}[ht]
\centering
\includegraphics[width=1.0\linewidth]{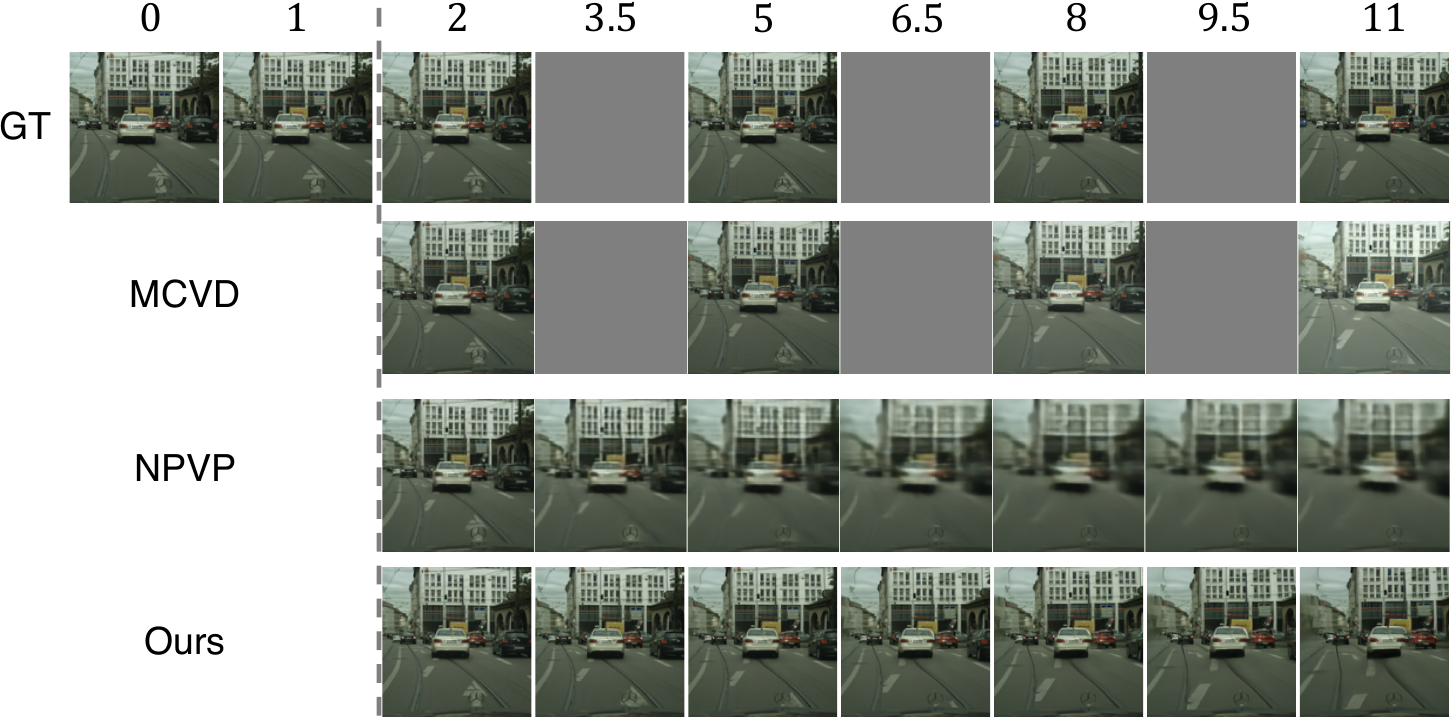}
\caption{Prediction examples on the Cityscapes dataset for MCVD \cite{voleti2022}, NPVP \cite{ye2023}, and our model. Gray frames indicate non-existent or unpredictable frames. MCVD exhibits issues with brightness changes and lacks the ability of continuous predictions. NPVP predictions are noticeably more blurry than the other two models.}
\label{fig:City_examples}
\end{figure}

The datasets presented in Table \ref{tab:KTH_CityScapes_KITTI_VFP_results} contains less stochasticity compared with SMMNIST and BAIR. For the KITTI dataset, STDiff outperforms previous work with a significant margin in terms of LPIPS and FVD. STDiff also achieves the best FVD score and second-best LPIPS score on the Cityscapes dataset. Some prediction examples of Cityscapes dataset are shown in Figure \ref{fig:City_examples}. Our model avoids the brightness changing issue seen in MCVD, and our predictions are sharper and more realistic compared to the NPVP model.

In general, the evaluation results show that STDiff has a better performance than previous deterministic and stochastic models in terms of either FVD or LPIPS. It is desirable because it is well known that the quality assessments produced by FVD and LPIPS are more aligned with human assessment than SSIM or PSNR \cite{unterthiner2019b, zhang2018}.

\subsection{Stochastic video prediction diversity}
\begin{table*}[ht]
\setlength{\tabcolsep}{1pt}
\centering
\begin{tabular}{lc|cccc} \hline
\multirow{3}{*}{Models} & \multirow{3}{*}{\#Parameters} & \multicolumn{2}{c}{KTH} & \multicolumn{2}{c}{SMMNIST}  \\
& & \multicolumn{2}{c}{\textit{10 $\rightarrow$ 20}} & \multicolumn{2}{c}{\textit{5 $\rightarrow$ 10}}\\
& & iLPIPS $\uparrow$ & FVD $\downarrow$ & iLPIPS $\uparrow$ & FVD $\downarrow$\\

\hline
NPVP \cite{ye2023} & 122.68M & 0.46 & 93.49 & 67.83 & 51.66 \\
Hier-VRNN \cite{castrejon2019} & 260.68M & 1.22 & 278.83 & 6.72 & 22.65\\
MCVD\cite{voleti2022} & 328.60M & \textbf{26.04} & 93.38 & \underline{123.93} & \underline{21.53}\\
STDiff-ODE & 201.91M & 15.60 & \underline{90.68} & 91.07 & 46.62  \\ \hline
\textit{STDiff} (ours) & 204.28M & \underline{25.08} & \textbf{89.67} & \textbf{136.27} & \textbf{14.93} \\
\hline

\end{tabular}

\caption{Stochastic video prediction diversity on KTH and SMMNIST datasets. A bigger iLPIPS denotes larger diversity, any deterministic prediction model has a iLPIPS of 0. Smaller FVD indicated better visual quality and more plausible predictions. \textbf{Boldface}: best results. \underline{Underlined}: second best results.}
\label{tab:iLPIPS_results}
\vspace{-3mm}
\end{table*}

In \citet{wang2022b}, generated images diversity is evaluated as LPIPS distance between different generated images. A greater LPIPS value between two generated images indicates increased dissimilarity in terms of content and structure. Likewise, we quantify the video diversity as average frame-by-frame LPIPS between different predicted video clips given the same past frames. To distinguish it from the standard LPIPS score between predictions and ground-truth, this diversity measure is termed as inter-LPIPS (iLPIPS). The iLPIPS results are listed in Table \ref{tab:iLPIPS_results}. All the iLPIPS values are reported at a $10^{-3}$ scale. Since iLPIPS does not incorporate the ground-truth, we use the Frechet Video Distance (FVD) as a supplementary metric. FVD ensures that randomly generated predictions exhibit satisfactory visual quality and that their distribution closely approximates the ground-truth distribution, i.e., the generated random predictions are plausible.

For all methods, we sampled 10 different random predictions for each test example to calculate the iLPIPS score. For KTH, we evaluated on the whole test set. For SMMNIST, we evaluated the results on 256 different test examples, similarly to MCVD \cite{voleti2022}. Besides, we take the same number of reverse diffusion sampling steps as MCVD, which is 100. Increasing the diffusion sampling steps would further improve the performance. NPVP \cite{ye2023} and Hier-VRNN \cite{castrejon2019} are trained based on their official code and MCVD \cite{voleti2022} is tested with their officially released trained models.

\begin{figure}[h]
\centering
\includegraphics[width=1.0\linewidth]{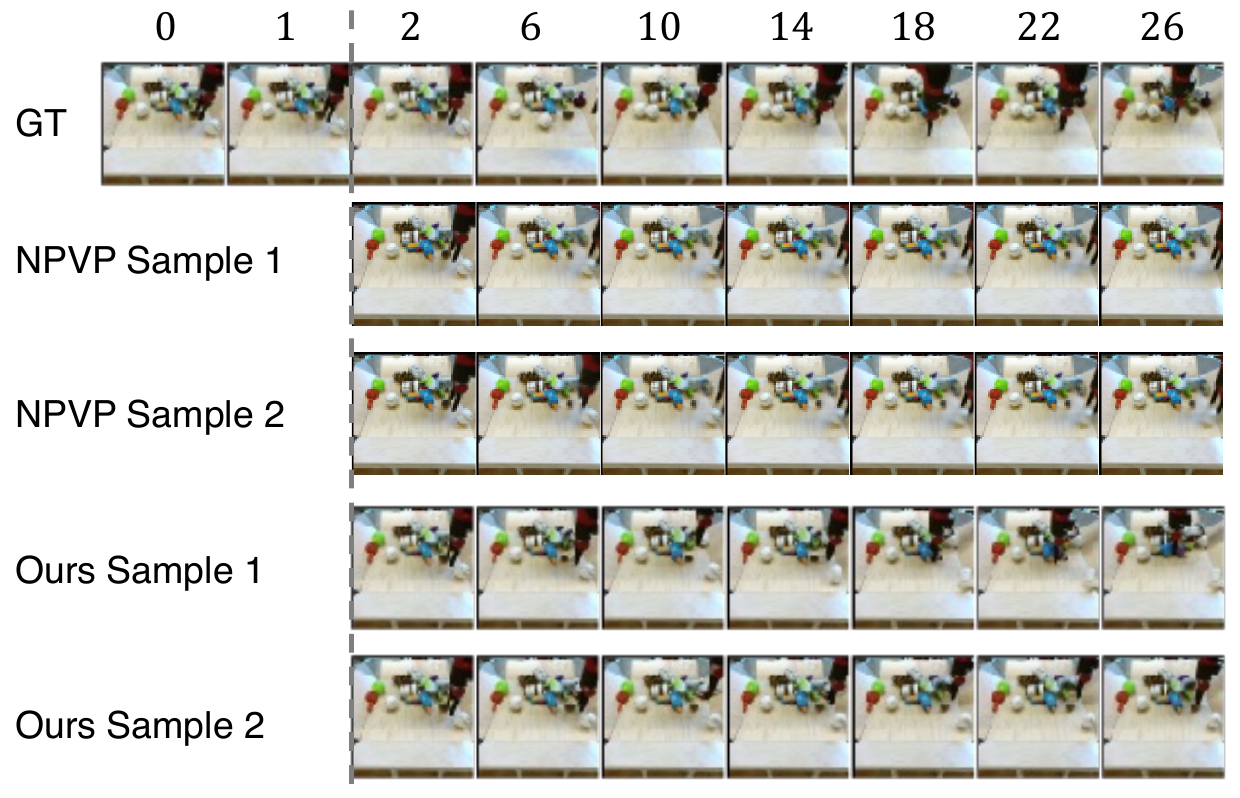}

\caption{Random predictions for BAIR dataset by our model and NPVP \cite{ye2023}.}
\label{fig:BAIR_examples}
\end{figure}

Comparing with the neural process-based NPVP method, The iLPIPS score of our STDiff model is more than 50 times bigger on KTH and twice bigger on SMMNIST. NPVP has a SOTA frame-by-frame SSIM or LPIPS performance, but the generated future frames lack diversity as assessed by iLPIPS. A visual examination of the results of NPVP validates this. We believe this is explained by the fact that NPVP only uses a single layer of latent variable for the VAE and uses a single global latent variable to account for the randomness of the whole sequence. Thus, NPVP has a very limited expressiveness for stochastic modeling. Figure \ref{fig:BAIR_examples} presents several uncurated predicted examples, demonstrating the better plausibility and greater diversity of our predictions compared to NPVP. Notably, NPVP tends to move the robot arm outside the field of view to minimize MSE.

\begin{table*}[ht]
\setlength{\tabcolsep}{1pt}
\centering

\begin{tabular}{lcccc|cccc} \hline
\multirow{3}{*}{Models} & \multicolumn{4}{c}{BAIR} & \multicolumn{4}{c}{KITTI}\\
& \multicolumn{4}{c}{\textit{2 $\rightarrow$ 28}, $2\times fps$} & \multicolumn{4}{c}{\textit{4 $\rightarrow$ 5}, $2\times fps$}\\
& FVD$\downarrow$ & SSIM$\uparrow$ & LPIPS$\downarrow$ &iLPIPS $\uparrow$ & FVD$\downarrow$ & SSIM$\uparrow$ & LPIPS$\downarrow$ &iLPIPS $\uparrow$ \\ 
\hline
Vid-ODE \cite{park2021b}& 2948.82 & 0.310& 322.58& 0 & 615.98 & 0.23 & 591.54 & 0\\
NPVP \cite{ye2023} & \underline{1159.14} & \underline{0.795} & \textbf{58.71} & \underline{1.06} & \underline{248.82} & \textbf{0.56} & \underline{313.94} & \underline{6.08}\\
\hline
\textit{STDiff} (ours)& \textbf{122.85} & \textbf{0.808} & \underline{75.37} & \textbf{109.33} & \textbf{78.39} & \underline{0.47} & \textbf{160.92} & \textbf{136.94}\\
\hline
\end{tabular}

\caption{Continuous VFP results on BAIR and KITTI. \textbf{Boldface}: best results. \underline{Underlined}: second best results.}
\label{tab:continuous_VFP}
\end{table*}

We also compare our model with Hier-VRNN that has 10 layers of latent variables. STDiff outperforms Hier-VRNN by a large margin on both datasets, despite Hier-VRNN having a larger model size. Compared with the SOTA diffusion-based model MCVD, STDiff outperforms it on SMMNIST and obtains a comparable iLPIPS for KTH. STDiff accomplishes this with a smaller model that has 100M less parameters. We believe that MCVD is not as efficient as STDiff for stochastic prediction because it learns the randomness of a whole video clip at once, similarly to NPVP, which requires a bigger model and also ignores an explicit temporal stochasticity learning. These results indicate that the good performance of our proposed method is not only attributed to the powerful image diffusion model, but also from our novel SDE-based recurrent motion predictor. 

In order to further investigate the effectiveness of our stochastic motion predictor, we implemented an ODE version of STDiff, which has the same architecture as STDiff, except that the motion predictor is ODE-based instead of SDE-based. We observe that STDiff has almost 1.5 times more diversity than STDiff-ODE on both datasets in terms of iLPIPS. Notably, our STDiff also achieves the best FVD on both datasets, highlighting that the prediction of STDiff, thanks to the use of an SDE, has large diversity and good visual quality simultaneously.

We can draw the conclusion that our STDiff has a better stochasticity modeling performance than previous stochastic models. The comparison with NPVP and Hier-VRNN validates the motivation that increasing the layers of latent variable for Hierarchical VAE is beneficial. And the comparison with STDiff-ODE and MCVD experimentally proves our claim that explicitly temporal stochasticity learning is also critical for a better diversity in future predictions. 

\subsection{Continuous prediction}

We summarize the continuous prediction performance of three continuous prediction models in Table \ref{tab:continuous_VFP}. MCVD is not included because it cannot conduct temporal continuous prediction. For this evaluation, we downsampled two datasets to 0.5 frame rate for training, then make the models predict with $2\times$ frame rate during test. This way, we get access to ground-truth high-frame rate test videos for metric calculations. The stochastic NPVP and STDiff predict 10 different random trajectories for each test example. In addition to iLPIPS and FVD, we also use the traditional evaluation protocol to report the best SSIM and LPIPS scores out of 10 different random predictions. 

STDiff outperforms Vid-ODE by a large margin in terms of all metrics. In addition to be deterministic and not decomposing the motion and content as we do, the capacity of Vid-ODE is too small. Indeed, performance of Vid-ODE on the original KTH dataset is not bad \cite{park2021b}, but downsampling largely increases the difficulty as the model needs to predict motion with a larger temporal gap. For the big, realistic and high resolution KITTI dataset, Vid-ODE fails to achieve reasonable performance mainly due to its limited model size. Examination of visual examples on BAIR shows that Vid-ODE cannot predict the stochastic motion of the robot arm and the image quality quickly degrades because of a large accumulated error.

As shown in Table \ref{tab:continuous_VFP}, for both datasets, STDiff achieves comparable or better performance in terms of SSIM and LPIPS compared to NPVP. However, we observe a significant performance gap in terms of FVD and iLPIPS with STDiff performing much better, especially for the dataset with more stochasticity, i.e., BAIR. This indicates that the capacity of randomness modeling also influences the performance of continuous prediction. Continuous prediction example on Cityscapes are shown in Figure \ref{fig:City_examples}, STDiff is able to predict frames at non-existent training dataset coordinates (e.g., 3.5, 6.5, and 9.5). Remarkably, STDiff holds the theoretical potential to predict videos at arbitrary frame rates.

\section{Conclusion}

In this paper, we propose a novel temporal continuous stochastic video prediction model. Specifically, we model both the spatial and temporal generative process as SDEs by integrating a SDE-based temporal motion predictor with a recurrent diffusion predictor, which greatly increases the stochastic expressiveness and also enables temporal continuous prediction. In this way, our model is able to predict future frames with an arbitrary frame rate and greater diversity. Our model reaches the SOTA in terms of FVD, LPIPS, and iLPIPS on multiple datasets.

\section*{Acknowledgements}
We thank FRQ-NT and REPARTI for the support of this research via the strategic cluster program.

\bibliography{aaai24}

\clearpage
\begin{center}
\textbf{\large Appendix}
\end{center}

\setcounter{equation}{0}
\setcounter{figure}{0}
\setcounter{table}{0}
\setcounter{page}{1}
\makeatletter
\renewcommand{\theequation}{S\arabic{equation}}
\renewcommand{\thefigure}{S\arabic{figure}}
\renewcommand{\thetable}{S\arabic{table}}

\maketitle
\section{Additional comparison experiments}

In response to reviewer requests, we compare STDiff with three more recent works on multiple datasets. Due to divergent experimental setups in these papers and the space constraints of the paper, we succinctly present the results below, employing a similar experimental protocol for fair comparison.

\vspace{-3mm}
\begin{table}[h]
\setlength{\tabcolsep}{1pt}
\centering
\begin{tabular}{l|cc} \hline
\multirow{2}{*}{Models} & \multicolumn{2}{c}{Cityscapes}\\
& SSIM $\uparrow$ & LPIPS $\downarrow$\\
\hline
DMVFN \cite{hu2023b} & 0.83 & 148.2 \\ \hline

\textit{STDiff} (ours) & \underline{0.80} & \textbf{52.68} \\
\hline

\end{tabular}
\label{tab:DMVF}
\vspace{-3mm}
\end{table}

\vspace{-3mm}
\begin{table}[h]
\setlength{\tabcolsep}{1pt}
\centering
\begin{tabular}{lc|ccc} \hline
\multirow{2}{*}{Models} & \multirow{2}{*}{\#Parameters} & \multicolumn{2}{c}{KTH} & BAIR\\
& & SSIM $\uparrow$ & LPIPS $\downarrow$ & FVD $\downarrow$\\
\hline
MOSO \cite{sun2023b} & 858M & 0.82 & 83.0 & 83.6 \\ \hline

\textit{STDiff} (ours) & 204.28M & \textbf{0.88} & \textbf{65.78} & \textbf{71.47} \\
\hline

\end{tabular}
\label{tab:MOSO}
\vspace{-3mm}
\end{table}

\vspace{-3mm}
\begin{table}[h]
\setlength{\tabcolsep}{1pt}
\centering
\begin{tabular}{l|cccc} \hline
\multirow{2}{*}{Models} & \multicolumn{3}{c}{KTH}\\
& SSIM $\uparrow$ & FVD $\downarrow$ & LPIPS $\downarrow$\\

\hline
MMVP \cite{zhong2023} & 0.91 & 424.25 & 239.25\\ \hline

\textit{STDiff} (ours) & 0.88 & \textbf{89.67} & \textbf{65.78} \\
\hline

\end{tabular}
\label{tab:MMVP}
\vspace{-3mm}
\end{table}

Our claim that \textit{STDiff} has the SOTA FVD and LPIPS score still holds when compared with all three papers. Additionally, \textit{STDiff} exhibits significantly higher efficiency compared to MOSO.

\section{Derivation of loss function}

To achieve conciseness, we establish our loss function based on the framework of denoising diffusion models \cite{sohl-dickstein2015, ho2020}. 

\noindent\textbf{Preliminaries.}
Given a noise schedule $\{\beta^l\in(0,1)\}_{l=1}^L$, where $l$ denotes the spatial diffusion step, the forward spatial diffusion process is defined as:
$$q(x^l|x^{l-1}) = \mathcal{N}(x^l;\sqrt{1-\beta^l}x^{l-1}, \beta^l\mathbf{I}).$$ 

The diffusion kernel can be derived using reparameterization trick as:
$$q(x^l|x^0) = \mathcal{N}(x^l;\sqrt{\bar{\alpha}^l}x^0, (1-\bar{\alpha}^l)\mathbf{I}),$$ where $\alpha^l=1-\beta^l$ and $\bar{\alpha}^l=\prod_{i=1}^l\alpha^i$. Then, we can directly sample $x^l$ by $x^l = \sqrt{\bar{\alpha}^l}x^0 + \sqrt{1-\bar{\alpha}^l}\epsilon^l$, where $\epsilon^l\sim\mathcal{N}(0,\mathbf{I})$. For each frame of the video, our target is to learn a model $p_\theta$ to reverse the forward spatial diffusion process conditioning on the motion feature and previous frame.

\noindent\textbf{Derivation.} Considering the conditional distribution 

\begin{align}
    p(\boldsymbol{x}|\boldsymbol{c}) &= p(\boldsymbol{x}|x^0_{t_0}, m_{t_0}) \nonumber\\ 
    &= \prod_{i=1}^{P}p(x_{t_i}^0|x^0_{t_{i-1}}, m_{t_{i-1}}) \nonumber\\
    &= \prod_{i=1}^{P}p(x_{t_i}^0|x^0_{t_{i-1}}, m_{t_{i}}). \nonumber
\end{align}

\noindent Note that motion feature $m_{t_i}$ is separately learned by the temporal diffusion process, it follows the transitional distribution $p(m_{t_i}|m_{t_{i-1}})$ and it is independent to $x_{t_i}$. As maximizing the log-likelihood is equivalent to minimizing the cross entropy between $p_\theta$ and data distribution $q$. In other words, the loss function can be defined as: 

\begin{align}
\mathcal{L}_{CE} &= E_{q}[-\log \prod_{i=1}^{P}p_\theta(x_{t_i}^0|x^0_{t_{i-1}}, m_{t_i})] \nonumber\\
&= \sum_{i=1}^{P}E_{q(x^0_{t_i})}[-\log p_\theta(x_{t_i}^0|x^0_{t_{i-1}}, m_{t_i})] \nonumber \\
&= \sum_{i=1}^{P}E_{q(x^0_{t_i})}[-\log (E_{q(x^{1:L}_{t_i}|x_{t_i}^0)}\frac{p_\theta(x_{t_i}^{0:L}|x^0_{t_{i-1}}, m_{t_i})}{q(x^{1:L}_{t_i}|x_{t_i}^0)})] \label{eq:importance_sampling} \\
&\leq \sum_{i=1}^{P}E_{q(x^0_{t_i})q(x^{1:L}_{t_i}|x_{t_i}^0)}[-\log \frac{p_\theta(x_{t_i}^{0:L}|x^0_{t_{i-1}}, m_{t_i})}{q(x^{1:L}_{t_i}|x_{t_i}^0)}] \label{eq:jensen},
\end{align}

where $q(x^{1:L}_{t_i}|x_{t_i}^0)$ denotes the joint distribution of all noised frames generated by the forward spatial diffusion process, and it is introduced by importance sampling in Eq. (\ref{eq:importance_sampling}). Jensen's inequality is applied from Eq. (\ref{eq:importance_sampling}) to Eq. (\ref{eq:jensen}). We can further expand Eq. (\ref{eq:jensen}) using the technique from Appendix B in \cite{sohl-dickstein2015} to be:

\begin{align}
\mathcal{L}_\theta = &\sum_{i=1}^P E_q[\log \frac{q(x_{t_i}^L|x_{t_i}^0)}{p_\theta(x_{t_i}^L)} \nonumber\\ 
&+ \sum_{l=2}^L\log \frac{q(x_{t_i}^{l-1}|x_{t_i}^l, 
x_{t_i}^0)}{p_\theta(x_{t_i}^{l-1}|x_{t_i}^l, x_{t_{i-1}}^0, m_{t_i})} \nonumber\\
&-\log p_\theta(x_{t_i}^0|x_{t_i}^1, x_{t_{i-1}}^0, m_{t_i})] \nonumber\\
= &\sum_{i=1}^P E_q[D_{KL}(q(x_{t_i}^L|x_{t_i}^0)||p_\theta(x_{t_i}^L)) \nonumber\\
&+\sum_{l=2}^{L}D_{KL}(q(x_{t_i}^{l-1}|x_{t_i}^l, 
x_{t_i}^0)||p_\theta(x_{t_i}^{l-1}|x_{t_i}^l, x_{t_{i-1}}^0, m_{t_i})) \nonumber\\
&-\log p_\theta(x_{t_i}^0|x_{t_i}^1, x_{t_{i-1}}^0, m_{t_i}) \nonumber\\ 
=&\sum_{i=1}^{P}\mathbb{E}_q[\mathcal{L}_L + \sum_{l>1}\mathcal{L}_{l-1} + \mathcal{L}_0] \label{eq:full_loss} 
\end{align}

\noindent In Eq. \ref{eq:full_loss}, $\mathcal{L}_L$ can be ignored since $p_\theta(x_{t_i}^L)$ follows a standard Gaussian distribution and there is no learnable parameter in $q$. $\mathcal{L}_0$ denotes the last spatial reverse diffusion step and it can also be ignored for a simplified loss function as in \cite{ho2020}. 

In $\mathcal{L}_{l-1}$, the reverse conditional posterior distribution $q(x_{t_i}^{l-1}|x_{t_i}^l,x_{t_i}^0)= \mathcal{N}(x_{t_i}^{l-1};\tilde{\mu}^l(x_{t_i}^l,x_{t_i}^0), \tilde{\beta}^{l}\mathbf{I})$ is tractable and can be derived using Bayes' rule. Where 

\begin{align}
\tilde{\mu}^l&=\frac{\sqrt{\bar{\alpha}^{l-1}}\beta^l}{1-\bar{\alpha}^l}x_{t_i}^0 + \frac{\sqrt{\alpha^l}(1-\bar{\alpha}^{l-1})}{1-\bar{\alpha}^l}x_{t_i}^l \nonumber\\
&= \frac{1}{\sqrt{\alpha^l}}(x_{t_i}^l - \frac{\beta^l}{\sqrt{1-\bar{\alpha}^l}}\epsilon^l)
\end{align}

\noindent and $\tilde{\beta}^l=\frac{1-\bar{\alpha}^{l-1}}{1-\bar{\alpha}^l}\beta^l$. Please see \cite{ho2020} for more details. 

And $p_\theta(x_{t_i}^{l-1}|x_{t_i}^l, x_{t_{i-1}}^0, m_{t_i}))$ also follows a Gaussian distribution $\mathcal{N}(\mu_\theta(x_{t_i}^l, l, x_{t_{i-1}}^0, m_{t_i}), \sigma^l\mathbf{I})$, where $\mu_\theta$ and $\sigma^l$ denote the mean and variance respectively. Similar to \cite{ho2020}, $\mu_\theta$ can be parameterized by a noise prediction UNet, i.e.,
\begin{align}
    \mu_\theta = \frac{1}{\sqrt{\alpha^l}}(x_{t_i}^l - \frac{\beta^l}{\sqrt{1-\bar{\alpha}^l}}\epsilon_\theta(x_{t_i}^l, x^0_{t_{i-1}}, m_{t_i}, l)).
\end{align}

Given these two Gaussian distributions, the KL divergence in $\mathcal{L}_{l-1}$ can be solved analytically:
\begin{align}
\mathcal{L}_{l-1} &\coloneqq \nonumber\\
&E_{x_{t_i}^0, \epsilon}[\frac{(1-\alpha^l)^2}{2(\sigma^l)^2\alpha^l(1-\bar{\alpha}^l)}\lVert\epsilon^l - \epsilon_\theta(x_{t_i}^l, x^0_{t_{i-1}}, m_{t_i}, l)\rVert]. \label{eq:complex_loss}
\end{align}

By substituting Eq. \ref{eq:complex_loss} into Eq. \ref{eq:full_loss} while ignoring the weighting term, akin to the methodology in \cite{ho2020}, we arrive at the final simplified loss function:

\begin{equation}
\mathcal{L}_\theta \coloneqq \sum_{i=1}^{P}E_{l}E_{x_{t_i}^0}E_{\epsilon\sim \mathcal{N}(0,1)}\left\lVert \epsilon^l - \bm{\epsilon}_\theta(x_{t_i}^l, x^0_{t_{i-1}}, m_{t_i}, l)\right\rVert^2.
\end{equation}

\section{Implementation details}

\subsection{Neural network architecture}
\noindent\textbf{Past frame motion learning} The motion feature extractor is a small CNN with 4 Conv-ReLU-MaxPool blocks, which takes the difference images as input and outputs a hidden feature for Conv-GRU. Conv-GRU takes the previous motion feature and the hidden  feature 
 as the inputs to generate the motion feature at the current time step. We implement the gates function in GRU as a one layer convolutional neural network. 

\noindent\textbf{Neural SDE} A small U-Net with two output heads are used to parameterize the $f_\theta$ for the neural SDE. We implement the SDE solver based on \textit{torchsde} project: \url{https://github.com/google-research/torchsde}. For SDE solver, we take "euler\_heun" method, with "$dt=0.1$". Empirically, we find that "stratonovich" SDE type is more stable than "ito" SDE type during training. The noise type for SDE is "diagonal". For the ODE solver, we take "euler" method with a step size of 0.1.

\noindent\textbf{Diffusion predictor} We implement the diffusion predictor based on the conditional image diffusion model from \textit{diffusers}: \url{https://huggingface.co/docs/diffusers/index}, with customized SPADE \cite{park2019} conditioning method to fuse the motion feature into the UNet. The previous clean frame is concatenated with the noisy current frame as the input of the UNet.

Please refer to the source code for more details about all the neural network architecture.

\subsection{Training details}

We utilized the AdamW optimizer with a learning rate of $1e^{-4}$ and employed a cosine annealing learning rate scheduler with warm restarts. The restart cycle was set to 200 epochs. To enhance the performance of the diffusion UNet, we applied exponential moving average. STDiff was trained for 600 epochs on all datasets.

To ensure the generalization of the neural SDE to unseen temporal coordinates during testing, we randomly sampled future frames during training. For the KTH dataset, the standard training procedure followed a $10\rightarrow 10$ scheme, where 10 past frames were used to predict 10 future frames. During training, we randomly sampled 6 future frames to predict ($P = 6$). The same sampling approach was applied to BAIR ($2\rightarrow10$), Cityscapes ($2\rightarrow10$), and SMMNIST ($5\rightarrow10$) datasets. For KITTI ($4\rightarrow 5$), we randomly sampled 2 future frames for prediction during training.

\subsection{inter-LPIPS (iLPIPS) calculation}

iLPIPS is firstly used in \citet{wang2022b} for generated images diversity evaluation. Here, we describe the detailed calculation of iLPIPS for generated video diversity measurement.

Let us denote a stochastic predicted video set as $\bm{\hat{V}}\in \mathbb{R}^{N\times S\times T \times H \times W \times C}$, where $N$ denotes the number of test examples, $S$ denotes the number of random predictions for each test example, and $T, H, W, C$ denote the video length, height, width and channels respectively. iLPIPS is calculated as

\begin{align}
    &iLPIPS = \nonumber \\
    &\frac{1}{U}
    \sum_{n=0}^{N-1} 
    \sum_{t=0}^{T-1}(\sum_{s=0}^{S-2}  \rho(\bm{\hat{V}}_{n, s, t}, \hat{\bm{V}}_{n, s+1, t}) + \rho(\bm{\hat{V}}_{n, 0, t}, \hat{\bm{V}}_{n, S-1, t})),
\end{align}
\noindent where $\rho()$ corresponds to LPIPS and $U=NST$. $iLPIPS\in[0,1]$ and a bigger value indicates a larger diversity. In summary, for $S$ predicted future video clip given the same past video clip, we take $S$ pairs of random video clips, and calculate the average frame-by-frame LPIPS score of each pair. While there are theoretically $C_2^S$ distinct pairs for $S$ different video clips, empirical results show that utilizing random $S$ pairs suffices for accurate reporting, and most importantly, it is much more computational efficient.

\section{Qualitative examples}
To assess the prediction quality and diversity of STDiff, we provided numerous example predicted videos by STDiff, as well as video examples generated by NPVP \cite{ye2023} and MCVD \cite{voleti2022} for visual quality comparison. The top left corner of each example video displays the temporal coordinates, with white coordinates representing past temporal points and red coordinates representing future temporal points. In addition to the submitted video examples, we invite readers to visit our project page at \url{https://github.com/XiYe20/STDiffProject} for online visualization of the results.

\end{document}